\documentclass[letterpaper, 10 pt, conference]{ieeeconf}  

\IEEEoverridecommandlockouts                              

\usepackage{graphicx}
\usepackage{multirow}
\usepackage{amsmath} 
\usepackage{amssymb}  
\usepackage[colorlinks=true,
            linkcolor=blue,
            urlcolor=blue,
            citecolor=green]{hyperref}

\DeclareMathOperator*{\argmin}{arg\,min}

\usepackage{xcolor}

\title{\LARGE \bf
PTZ-Calib: Robust Pan-Tilt-Zoom Camera Calibration
}

\author{Jinhui Guo$^{1}$, Lubin Fan$^{1\dagger}$, Bojian Wu$^{2}$, Jiaqi Gu$^{1}$, Shen Cao$^{1}$ and Jieping Ye$^{1}$
\thanks{$^{1}$ Alibaba Cloud Computing; $^{2}$ Independent Researcher}
\thanks{$\dagger$ Corresponding author. Email: lubin.flb@alibaba-inc.com.}
}

\begin{document}

\maketitle
\thispagestyle{empty}
\pagestyle{empty}

\begin{abstract}
In this paper, we present \emph{PTZ-Calib}, a robust two-stage PTZ camera calibration method, that efficiently and accurately estimates camera parameters for arbitrary viewpoints. Our method includes an offline and an online stage. In the offline stage, we first uniformly select a set of reference images that sufficiently overlap to encompass a complete $360^{\circ}$ view. We then utilize the novel PTZ-IBA (PTZ Incremental Bundle Adjustment) algorithm to automatically calibrate the cameras within a local coordinate system. Additionally, for practical application, we can further optimize camera parameters and align them with the geographic coordinate system using extra global reference 3D information. In the online stage, we formulate the calibration of any new viewpoints as a relocalization problem. Our approach balances the accuracy and computational efficiency to meet real-world demands. Extensive evaluations demonstrate our robustness and superior performance over state-of-the-art methods on various real and synthetic datasets. Datasets and source code can be accessed online at \url{https://github.com/gjgjh/PTZ-Calib}
\end{abstract}

\section{Introduction}

Pan-Tilt-Zoom (PTZ) cameras are widely used in video broadcasting~\cite{Liu:2023:LAC,homayounfar2017sports} and security surveillance~\cite{Wu:2013:KPTZ, Eldrandaly:2019:Coverage}. For applications, like panorama creation, localization or digital twins ~\cite{lalonde2007system,rameau2013self,wu2012keeping,Song:2006:Traffic}, accurately estimating their orientation and intrinsic parameters is highly desirable.
However, the camera parameters provided by the hardware contain multiple errors~\cite{Wu:2013:KPTZ}, which will impact the effectiveness of their applications. Therefore, it is necessary to propose an efficient and accurate calibration method for PTZ cameras.

Two types of PTZ camera calibration tasks are popular. One category focuses on sports field registration, which is usually equivalent to the homography estimation problem, achieved by local feature matching~\cite{homayounfar2017sports,Gupta:2011:ULE,Nie2021ARA,Sha:2020:EtE} or learning-based methods~\cite{jiang2020optimizing,SFLNet}.
While these methods are effective in specific applications, their assumption of planarity and known fixed structure restricts the use in complex 3D scenes. Another category addresses the more general task of PTZ camera calibration. To achieve efficient and precise estimation results, these methods often incorporate numerous assumptions to simplify the problem. For example, the camera has a known initial position~\cite{PTZ-SLAM}, rotates only~\cite{selfcalib1997}, lacks distortion~\cite{chen2018two}, or assumes simple illumination and environmental changes~\cite{Sinha:2006:Mosaic}. 

\begin{figure}[!t]
    \centering
    \includegraphics[width=\linewidth]{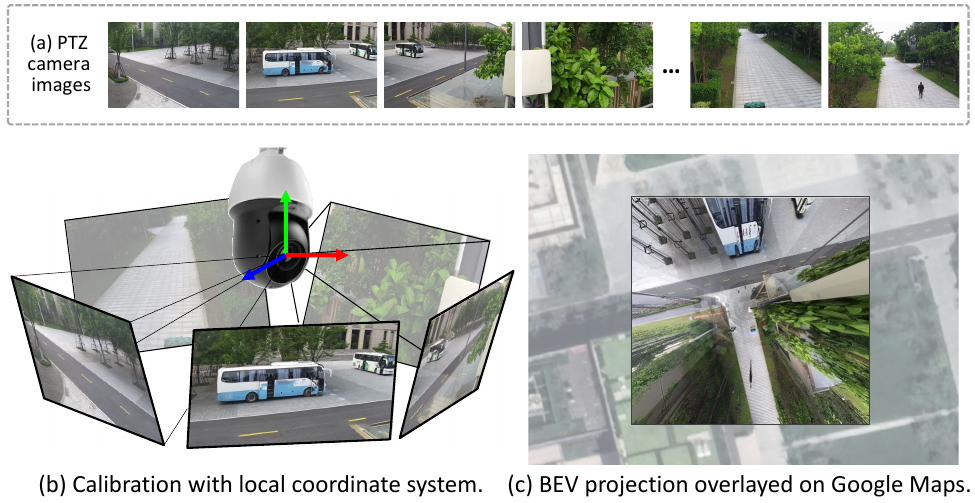}
    \caption{\textbf{Overview of PTZ-Calib}. Given a set of camera images(a), our method can automatically calibrate them in the local coordinate system(b) and further align them geographically using global 3D references(c).}
    \label{fig:teaser}
\end{figure}

To address these challenges, our proposed method robustly estimates camera parameters, such as camera pose, focal length, and distortion coefficients, for any viewpoint via a two-stage process. In the offline stage, we select uncalibrated reference frames and track 2D feature points across them using feature matching algorithm. We then apply our novel PTZ-IBA (PTZ Incremental Bundle Adjustment) algorithm for automatic calibration in local coordinate system. To further optimize camera parameters, we incorporate global geographic reference 3D data, aligning them with the geographic coordinate system.
In the online stage, we calibrate camera parameters for new viewpoints using feature matching and optimization techniques.
This method effectively balances accuracy and computational efficiency.

Extensive evaluations validate the efficacy of our method. In sports field registration, our algorithm outperforms leading open-sourced state-of-the-art methods and achieves the best results. We also build a set of 3D synthetic scenes to evaluate the calibration accuracy. Our method consistently delivers superior performance both quantitatively and qualitatively. Moreover, we demonstrate the application of our algorithm in real-world settings, illustrating the broad applicability in complex scenarios. We summarize our contributions as follows. 

\begin{itemize}
    \item We propose a robust two-stage PTZ camera calibration method \emph{PTZ-Calib} to provide efficient and accurate camera parameters for arbitrary viewpoints.
    \item We introduce the PTZ-IBA algorithm, which automatically calibrates camera views within a local coordinate system. Besides, we can further align parameters to real-world global coordinates using extra 3D information.
    \item Extensive evaluations demonstrate that our method outperforms current SOTA methods across diverse real and synthetic datasets, as well as various applications.
\end{itemize}

\section{Related Work}\label{sec:related}

\textbf{Sports Field Registration}
involves estimating correspondences (such as lines and circles) between a sports field and images captured by a fixed, rotatable camera, traditionally treated as a homography estimation problem. Early methods identified correspondences and obtained homography using Direct Linear Transform (DLT) or optimization-based method~\cite{homayounfar2017sports, Gupta:2011:ULE,Nie2021ARA}. Recent methods have shifted towards semantic segmentation to learn the sports field's representation, directly predicting or regressing an initial homograph matrix~\cite{jiang2020optimizing}, or searching for the optimal homography in a database of synthetic images with predefined matrices or known camera parameters~\cite{Sha:2020:EtE,Chen2019SportsCC}. End-to-end approaches~\cite{SFLNet} also obtained promising results.
Recently,~\cite{Theiner:2023:TVCalib,Marc:2024:NBJW} consider the task as a camera calibration problem, leveraging the segment correspondences to predict the camera pose and focal length. 
Despite achieving notable successes, the limitation lies in their inability to be readily adapted to alternative application scenarios due to the scarcity of training data and complex 3D scenes.

\textbf{PTZ Camera Calibration }
aims to estimate full camera parameters, including intrinsic and extrinsic parameters.
Earlier approaches often rely on various assumptions to simplify the problem. \cite{Wu:2013:KPTZ} develops a comprehensive model that includes a dynamic correction process to maintain accurate calibration over time. However, this model assumes a consistent relationship between zoom scale and focal length.
Other methods assume that cameras start with known initial camera poses~\cite{PTZ-SLAM}, rotate only~\cite{selfcalib1997} or have no lens distortions~\cite{chen2018two}, which is impractical in real-world applications. \cite{Liu:2023:LAC} presents a novel linear auto-calibration method for bullet-type PTZ cameras only and cannot directly apply to other cameras without modifications. 

In comparison, our method relies on minimal assumptions while fully calibrating the camera parameters. Moreover, with additional 2D-3D annotations, we can further improve the camera parameters and obtain absolute poses in the geographic coordinate system. By employing a two-stage design, our method effectively balances computational efficiency with accuracy, making it highly practical for various applications.

\section{Preliminary}
\label{sec:PRELIMINARY}

\subsection{Pinhole camera model}

We first review the imaging process of the pinhole camera model. Denote a point in 3D space with homogeneous coordinates as $\mathbf{X}= [X,Y,Z,1]^T$. The corresponding coordinates $\mathbf{X'}=[X',Y',Z',1]^T$ in the camera coordinate system can be calculated as follows:
\begin{equation}
\mathbf{X'} = \mathbf{R}[\mathbf{I}|-\mathbf{C}] \mathbf{X},
\end{equation}
where $\mathbf{R}$ denotes the rotation matrix in the world coordinate system, and $\mathbf{C}$ indicates the position of the camera's projection center. The projected 2D coordinates on the normalized imaging plane can be represented as follows:
\begin{equation}
\mathbf{x}' = [x',y',1]^T = [X'/Z',Y'/Z',1]^T.
\end{equation}

Considering that PTZ cameras often utilize wide-angle lenses, the image distortion is unavoidable. Therefore, we also model the distorted coordinates $\mathbf{x}''$ as follows,
\begin{equation}
\begin{aligned}
    x'' = {x'(1 + {k_1}{r^2} + {k_2}{r^4}) + 2{p_1}x'y' + {p_2}({r^2} + 2{{x'}^2})},\\
    y'' = {y'(1 + {k_1}{r^2} + {k_2}{r^4}) + {p_1}({r^2} + 2{{y'}^2}) + 2{p_2}x'y'},\\
\end{aligned}
\end{equation}
\noindent where $k_1$ and $k_2$ denote the radial distortion coefficients, while $p_1$ and $p_2$ represent the tangential distortion coefficients, with ${r^2} = {{x'}^2} + {{y'}^2}$. 

The coordinates of $\mathbf{x}$ on the image can be denoted by:
\begin{equation}
\mathbf{x} = \mathbf{K}\mathbf{x}'' 
= 
\begin{bmatrix}
    f & 0 & c_x \\
    0 & f & c_y \\
    0 & 0 & 1
\end{bmatrix}
\begin{bmatrix}
    x'' \\
    y'' \\
    1
\end{bmatrix},
\end{equation}
\noindent where $\mathbf{K}$ is the intrinsic matrix. For simplicity, we assume square pixels, a principle point at the image center and a zero skew factor in our model.

\subsection{PTZ camera model}

A PTZ camera, capable of directional and zoom control, is typically mounted in a fixed, elevated position. It means that the camera primarily experiences rotation-only motions~\cite{hayman2003effects}, without offset between the camera's projection center and its rotation center. In our settings, the estimated parameters include the focal length $f$, the distortion coefficients $k1$, $k2$, $p1$, $p2$, as well as the rotation matrix $\mathbf{R}$ for each camera view. The other parameters, such as the principal points $cx$, $cy$, camera position $\mathbf{C}$, remain fixed for the same PTZ camera.

\begin{figure*}[!t]
    \centering
    \includegraphics[width=\linewidth]{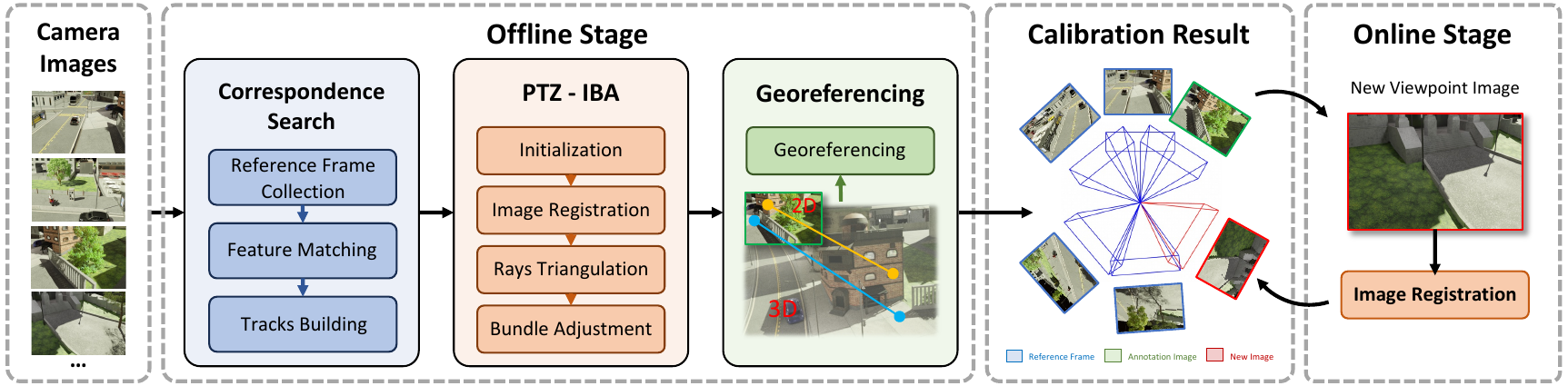}
    \caption{\textbf{Pipeline of the two-stage \emph{PTZ-Calib} method}. In the offline stage, the reference images selected from camera images are auto-calibrated with our novel \emph{PTZ-IBA} algorithm. Once online, our method estimates the camera parameters for arbitrary viewport images accurately and efficiently.}
    \label{fig:overview}
\end{figure*}

\subsection{Ray landmark} 

We define a ray as a directional vector originating from the camera's projection center. The \textit{ray landmark} is then represented by a normalized three-dimensional vector. In our method, the ray landmarks serve as the primary representation of scenes and constrain the relative rotation and intrinsic parameters of the camera across different views. However, relying solely on ray landmarks is insufficient for determining the absolute pose of camera in the real-world geographic coordinate system due to the lack of explicit 3D information. Therefore, incorporating 3D landmarks, which represent the 3D points in the world, becomes necessary.

\section{Method}

\subsection{Overview}

As illustrated in Fig.~\ref{fig:overview}, we propose a two-stage method, consisting of an offline stage and an online stage. The purpose of the offline stage is to estimate camera parameters of some reference frames to better serve the efficient arbitrary viewpoint calibration during the online stage. Specifically, in the offline stage, we select some reference frames based on certain rules, and establish track information of 2D feature points between reference frames using an image feature matching algorithm. Based on this, we propose the novel PTZ-IBA algorithm (PTZ Incremental Bundle Adjustment), which primarily focuses on constraining the consistency of ray landmark projections for all viewpoints and \textit{automatically} obtains camera parameters in the \textit{local} coordinate system. 
Finally, in order to build the correspondences between 2D images and 3D real-world coordinates, by incorporating a few 2D-3D annotations, we further fine-tune the camera parameters and obtain the \textit{global} absolute poses of the camera.

After completing the calibration of all reference frames in the offline stage, the online stage becomes very intuitive. Given an image from a new viewpoint, we can calibrate the corresponding camera parameters using the similar aforementioned feature matching and optimization methods. It's worth noting that, our two-stage approach divides the heavy computational tasks into the offline stage and keeps the lighter computations for the online stage, ensuring a balance between accuracy and computational efficiency.

\subsection{Offline stage}

\subsubsection{Correspondence search}
We collect reference images and build data associations with feature matching in this step.

\textbf{Reference Frame Collection.}
PTZ cameras inherently offer wide field of view (FOV) coverage. To ensure comprehensive scene representation through reference images, a 360-degree image collection is essential. In practice, we typically capture 8 to 10 images uniformly at the camera's minimum zoom level to ensure sufficient overlap.
Depending on the scene's complexity, we also collect more images at different zoom and tilt levels, resulting in a final set of about 30 to 50 reference images.

\textbf{Feature Matching.}
To tackle the challenging variations in image appearance caused by illumination, weather, and seasonal changes, etc., for robustness, we employ the state-of-the-art deep feature extraction and matching techniques, such as SuperPoint~\cite{detone2018superpoint} and SuperGlue~\cite{sarlin20superglue}, to create pixel-level correspondences among all reference images. Additionally, we use RANSAC in conjunction with homography transformation to filter out outliers from the matching results.

\textbf{Tracks Building.}
To strengthen the global constraints across reference frames and estimate more accurate camera parameters, we consolidate pairwise matching results to establish tracks. A track represents the matching relationships of keypoints across multiple images. In the PTZ camera model, each track corresponds to a ray landmark. We employ the Union-Find algorithm to build the tracks~\cite{moulon2012unordered} and remove any that are either too short or have matching conflicts.

\subsubsection{PTZ-IBA}

In this step, we aim to determine the relative poses and intrinsic parameters of the reference frames. 
Simplify, we build a local coordinate system with an initial frame as reference, the camera's projection center $\mathbf{C}$ is $\mathbf{0}$.

\textbf{Initialization.}
Choosing the initial frame and the next best view is critical~\cite{schoenberger2016sfm}. We select the frame with the highest number of neighborhood matches as the initial starting frame. If the incremental optimization from this initial frame fails, we then opt for the frame with the second-highest number of neighborhood matches as the new starting frame, and continue this process as needed. For selecting the next best view, we pick the frame that has the most neighborhood matches among the neighbors of the already registered frames as the next frame to be registered. This strategy is key to ensuring the effectiveness of the parameter estimation process.

The initial focal length can be empirically determined by $f=1.2*\max(W, H)$, where $W$ and $H$ are width and height of image. The initial distortion coefficients can be set to 0, and the initial relative rotation $\mathbf{R}_{ij}$ between the image pair can be calculated as $\mathbf{R}_{ij}=\mathbf{K}_i^{-1}\mathbf{H}_{ij}\mathbf{K}_j$, where $\mathbf{H}_{ij}$ is the homography matrix between the image $i$ and $j$.

\textbf{Image Registration.}
A new image can be registered by solving an optimization problem that uses feature matches with images that have already been registered. 
First, the ray landmarks of the reference registered image can be derived based on the pixel correspondences. This step is essentially the reverse of the imaging model used for the PTZ camera. Denote the undistorted point on the reference image as $\mathbf{x}$, the corresponding ray landmark $\mathbf{r}$ can be calculated as follows:
\begin{equation}
\label{equ:pix_to_ray}
\begin{aligned}
    \mathbf{r}' = \mathbf{R}^{-1} \mathbf{K}^{-1} \mathbf{x}, \mathbf{r} = \frac{\mathbf{r}'}{\|\mathbf{r}'\|}. \\
\end{aligned}
\end{equation}

We minimize the reprojection error of the ray landmarks on the current frame $i$ to estimate the camera parameters $\mathbf{P}_i$:
\begin{equation}
\label{equ:img_reg}
    \mathbf{P}_i^* = \argmin_{\mathbf{P}_i} \sum^m_k \rho \left( \| \pi(\mathbf{P}_i, \mathbf{r}_k) - \mathbf{x}_{ik} \|^2 \right),
\end{equation}

\noindent where $\pi(\cdot)$ denotes the function that projects ray landmarks to image space, as described in Section \ref{sec:PRELIMINARY}. $\rho(\cdot)$ is a robust loss function that mitigates the influence of outliers. $\mathbf{x}_{ik}$ represents the observation of ray landmark $\mathbf{r}_k$ on the current image $i$, while $m$ denotes the number of rays.

\textbf{Rays Triangulation.}
Given that images captured by the PTZ camera do not involve translation, traditional triangulation methods cannot directly determine the positions of 3D landmarks.
In rays triangulation, a ray landmark is typically observed by multiple frames. Using the parameters from the multi-frame viewpoints as inputs, 
individual ray coordinates can be computed using (\ref{equ:pix_to_ray}). The final estimated position is then determined by calculating the average of these ray coordinates.

\textbf{Bundle Adjustment.}
Image registration and rays triangulation usually yield only approximate estimates. As the number of images increases, global bundle adjustment becomes crucial for mitigating drift errors and maintaining global consistency.
Global bundle adjustment performs joint optimization on all currently registered images and ray landmarks to yield more precise estimation results. The goal is to minimize all reprojection errors:
\begin{equation}
    \mathbf{P}^*, \mathbf{r}^* = \argmin_{\mathbf{P}, \mathbf{r}} \sum^n_i \sum^m_k \rho \left( \| \pi(\mathbf{P}_i, \mathbf{r}_k) - \mathbf{x}_{ik} \|^2 \right),
\end{equation}
\noindent where $n$ is the number of registered images and $m$ is the number of rays.

While global bundle adjustment can achieve high-accuracy parameter estimates, it requires significant computational resources.
Therefore, global bundle adjustment is performed only after the number of registered images increases by a certain factor. The entire process is conducted once all images are registered. Any image that fails to register will be discarded after a certain number of attempts.

\subsubsection{Georeferencing}
PTZ-IBA only provides the pose and camera intrinsic parameters within a \textit{local} coordinate system. To fulfill the requirements for practical applications in urban environments, we need additional 2D-3D correspondences to align the PTZ camera with geographic coordinate systems like UTM. 
The 2D keypoints are annotated from the images, while the 3D keypoints come from oblique 3D models or Digital Orthophoto Map (DOM) obtained from drones. To reduce the manual annotation costs, we select only a small number of images (typically 2 to 4) for 2D-3D annotations.

Denote $\mathbf{T}^l_w \in \mathrm{SE}(3)$ as the transformation matrix from the geographical coordinate system to the local coordinate system. We minimize the following loss function to obtain $\mathbf{T}^l_w$ and the fine-tuned camera parameters:
\begin{equation}
\begin{aligned}
    \mathbf{P}^*, \mathbf{r}^*, {\mathbf{T}^l_w}^* &= \argmin_{\mathbf{P}, \mathbf{r}, \mathbf{T}^l_w} \sum^n_i \sum^{m_1}_k \rho \left( \| \pi(\mathbf{P}_i, \mathbf{r}_k) - \mathbf{x}_{ik} \|^2 \right) \\
    &+ \sum_{i \in \mathbf{S}} \sum^{m_2}_j \rho \left( \| \pi(\mathbf{P}_i, \mathbf{T}^l_w \mathbf{X}_j) - \mathbf{x}_{ij} \|^2 \right),
\end{aligned}
\end{equation}
\noindent where $\mathbf{S}$ represents the set of annotated images and $\mathbf{X}$ denotes the annotated 3D points. 
To clarify, here $m_1$ represents the number of rays, and $m_2$ indicates the number of 2D-3D correspondences. Once $\mathbf{T}^l_w$ is obtained, the absolute poses in the geographical coordinate system can be recovered. 

\subsection{Online stage}

During the online stage, we aim to estimate the camera parameters of arbitrary viewpoints efficiently and accurately. We model the online parameter estimation as a relocalization problem. Specifically, with a set of views calibrated in the offline stage, we determine the optimal camera parameters for the given view in order to best align with the correspondences. Initially, we match the current input image with the most recent image captured from the video stream, perform image registration, and estimate the initial camera parameters using (\ref{equ:img_reg}). Notably, the initial camera parameters may contain accumulated errors. We compute the frustum overlap between the current image and all the calibrated reference image based on the estimated initial parameters. 
We perform matching on images with an overlap greater than a threshold and select the best-matched image as the reference frame. Finally, the camera parameters for the current frame are obtained using (\ref{equ:img_reg}).
This substantially reduces the number of matches required, thereby speeding up the computation in the online stage as well as improving the accuracy.

\section{Experiments}

\begin{figure*}[!t]
    \centering
    \includegraphics[width=\linewidth]{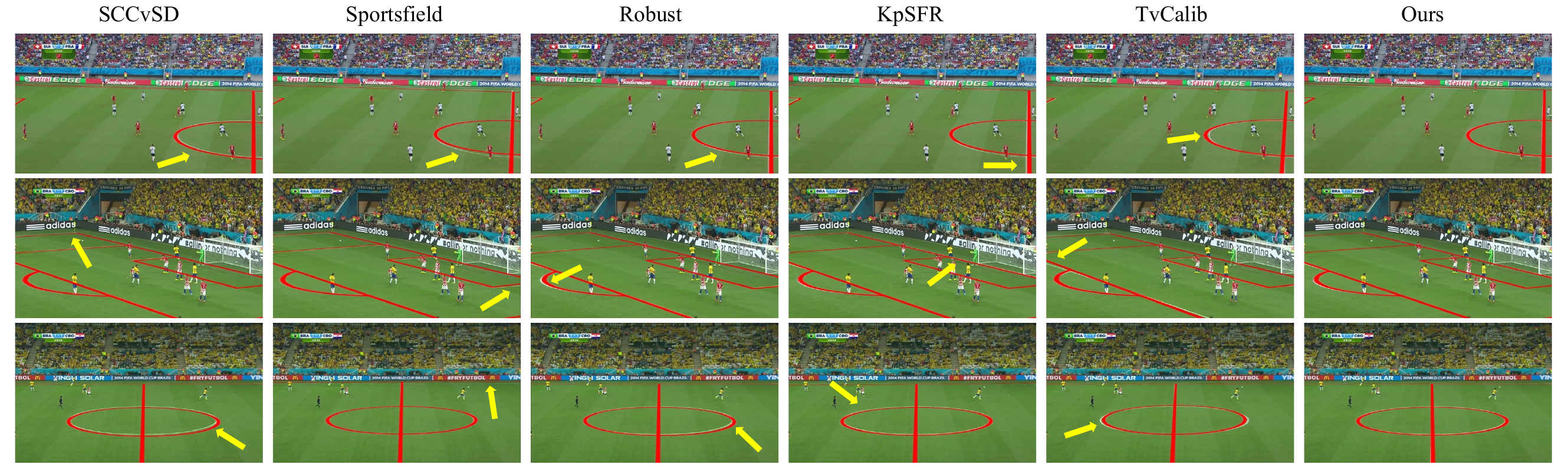}
    \caption{\textbf{Comparisons of sports field registration on the WorldCup dataset.} The sports field template is projected onto the image in red based on the estimated camera parameters. Misregistrations are pointed out with yellow arrows.}
    \label{fig:worldcup-compare}
\end{figure*}

\subsection{Experimental setup}

\textbf{WorldCup dataset~\cite{homayounfar2017sports}.}
This dataset, widely used for sports field registration, was collected during the 2014 World Cup held in Brazil. It includes 209 training images and 186 test images from different matches. Each image is manually annotated with a homography matrix and grass segmentation. Unlike typical methods that use the homography for transformations, our algorithm estimates the camera's intrinsic and extrinsic parameters to derive the geometric transformation.

Since our method requires a fixed camera position, we only use a subset, excluding matches like COL vs. IVC and ARG vs. IRN with insufficient coverage. We assume constant camera positions for matches at the same venue and conduct experiments accordingly, as detailed in Table~\ref{tab:subset_dataset}. Each training image is annotated with 30 random 2D-3D correspondences.

\begin{table}[ht]
\caption{Subset dataset used in experiments.}
\label{tab:subset_dataset}
    \centering
    \scalebox{0.75}{
    \begin{tabular}{c c c c c c}
       \hline
       \multirow{2}{*}{Split} & \multirow{2}{*}{Sports Venue} & \multicolumn{2}{c}{Train} & \multicolumn{2}{c}{Test} \\
       \cline{3-6}
        & & Match & Frame ID & Match & Frame ID \\
        \hline
        1 & Arena Corinthians & NED vs. ARG & 1-15 & ARG vs. SUI & 159-175 \\
        2 & Arena Corinthians & NED vs. ARG & 1-15 & BRA vs. CRO & 1-22 \\
        3 & Arena Corinthians & NED vs. ARG & 1-15 & URU vs. ENG & 85-96 \\
        4 & Arena Pernambuco & USA vs. GER & 120-134 & CRO vs. MEX & 126-142 \\
        5 & Maracanã Stadium & GER vs. ARG & 135-149 & ESP vs. CHI & 97-110 \\
        6 & Maracanã Stadium & GER vs. ARG & 135-149 & FRA vs. GER & 176-186 \\
        7 & Arena Fonte Nova & GER vs. POR & 170-191 & SUI vs. FRA & 111-125 \\
        \hline
    \end{tabular}
    }
\end{table}

As the WorldCup dataset lacks explicit camera parameters, we evaluate the accuracy using the Intersection over Union (IoU), divided into $IoU_{part}$ and $IoU_{whole}$. $IoU_{part}$ calculates the IoU for the camera's visible area projected onto the Bird's Eye View (BEV), using both predicted and ground truth parameters, but within the sports field template. For $IoU_{whole}$, we project the full visible area onto BEV using ground truth parameters, then re-project it onto the image with both predicted and ground truth parameters to compute the IoU between these projections.

\textbf{Synthetic dataset.}
Due to the lack of publicly available datasets for PTZ camera parameter estimation, we create a synthetic dataset using Blender for assessment. We design 10 datasets, split evenly for both indoor and outdoor scenes, each from a single PTZ camera with 3 random configurations of focal lengths and tilt angles. Each configuration involves a full rotation of the camera, producing 180 images that cover the scene. Each image provides ground truth camera intrinsic and extrinsic parameters. Additionally, we generate several 3D points around the camera and project them onto the images to form 2D-3D correspondences, with each image containing 30 pairs. To simulate real-world conditions, zero-mean Gaussian noise of 3.0 pixels was added to each projected point. We plan to release this dataset to the public.

For evaluating the intrinsic parameter, we use the absolute focal length error $FLE = |f - f'|$, where $f$ is the predicted and $f'$ is the ground truth. For extrinsic parameters, we employ the Absolute Pose Error metrics, specifically the rotation angle error $APE_{rot}$ and the translation error $APE_{trans}$.

\textbf{Implementation Details.} 
Our system is implemented in C++ on a CentOS platform, utilizing 2.50 GHz Intel CPU and NVIDIA T4 GPU. For feature matching, we employed fp16 quantization acceleration for SuperPoint and SuperGlue. The RANSAC geometric verification threshold is set to 4 pixels, with a minimum number of matching pairs set at 40. Non-linear optimization is implemented using the Ceres~\cite{Ceres}.

\subsection{Results}

\textbf{WorldCup dataset.}
We selected the open-sourced state-of-the-art sports field registration methods~\cite{Chen2019SportsCC,jiang2020optimizing,Nie2021ARA,Chu_2022_CVPR,Theiner:2023:TVCalib} for comparison. In our experiments, we utilized the pre-trained weights and default parameters for fairness.

\begin{figure}[ht]
    \centering
    \includegraphics[width=\linewidth]{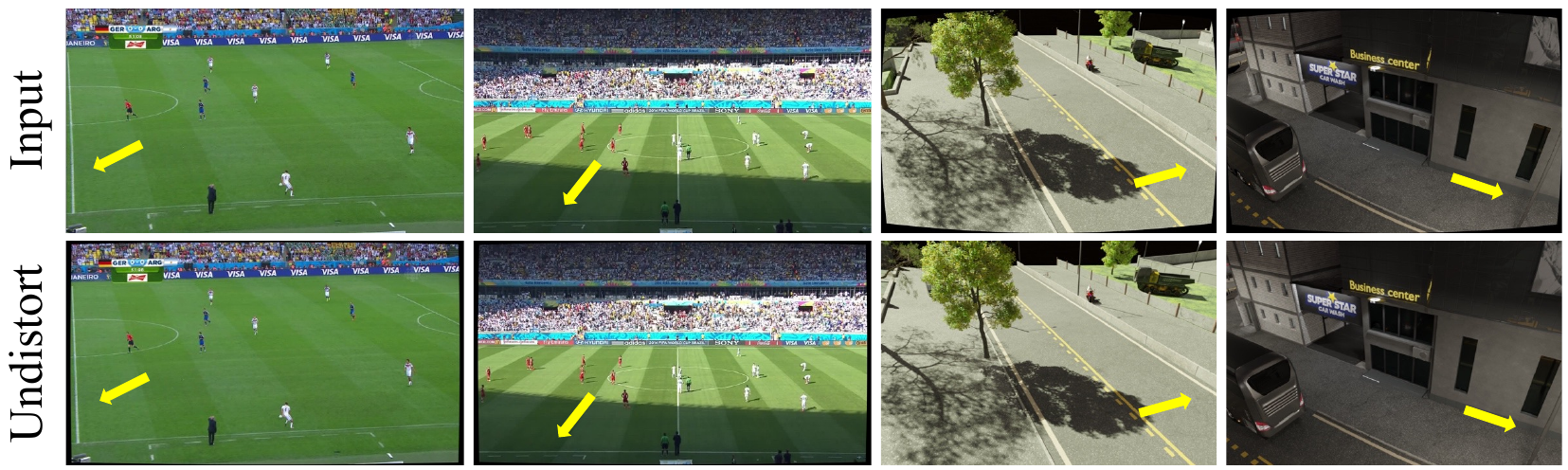}
    \caption{\textbf{Undistortion results.} Regions to focus on are highlighted.}
    \label{fig:all-undistort}
\end{figure}

As in Table~\ref{tab:Quantitative_WorldCup}, our method achieves the best performance. We projected the sports field template onto the 2D images for overlay visualization. As in Fig. \ref{fig:worldcup-compare}, our results estimate the best registration results compared with other methods. Undistortion corrections are visualized in Fig.~\ref{fig:all-undistort}, highlighting our effective estimation of distortion coefficients, which straightens field lines in the undistorted images.

\begin{table}[t!]
\caption{Comparisons of IoU on the WorldCup dataset.}
\label{tab:Quantitative_WorldCup}
    \centering
    \resizebox{0.63\linewidth}{!}{%
    \begin{tabular}{c c c c c}
       \hline
       \multirow{2}{*}{Method} & \multicolumn{2}{c}{$IoU_{part}\uparrow$} & \multicolumn{2}{c}{$IoU_{whole}\uparrow$} \\
       \cline{2-5}
        & mean & median & mean & median \\
        \hline
        SCCvSD\cite{Chen2019SportsCC} & 95.7 & 96.9 & 90.6 & 92.3 \\
        Sportsfield\cite{jiang2020optimizing} & 96.2 & 97.4 & 91.1 & 93.3 \\
        Robust\cite{Nie2021ARA} & 96.5 & 97.6 & 91.4 & 93.0 \\
        KpSFR\cite{Chu_2022_CVPR} & 96.7 & 97.6 & 92.5 & 94.2 \\
        TVCalib\cite{Theiner:2023:TVCalib} & 95.8 & 97.0 & 90.9 & 93.2 \\
        Ours & \textbf{96.8} & \textbf{97.8} & \textbf{93.1} & \textbf{94.4} \\
        \hline
    \end{tabular}%
    }
\end{table}

\begin{figure}[!t]
    \centering
    \includegraphics[width=\linewidth]{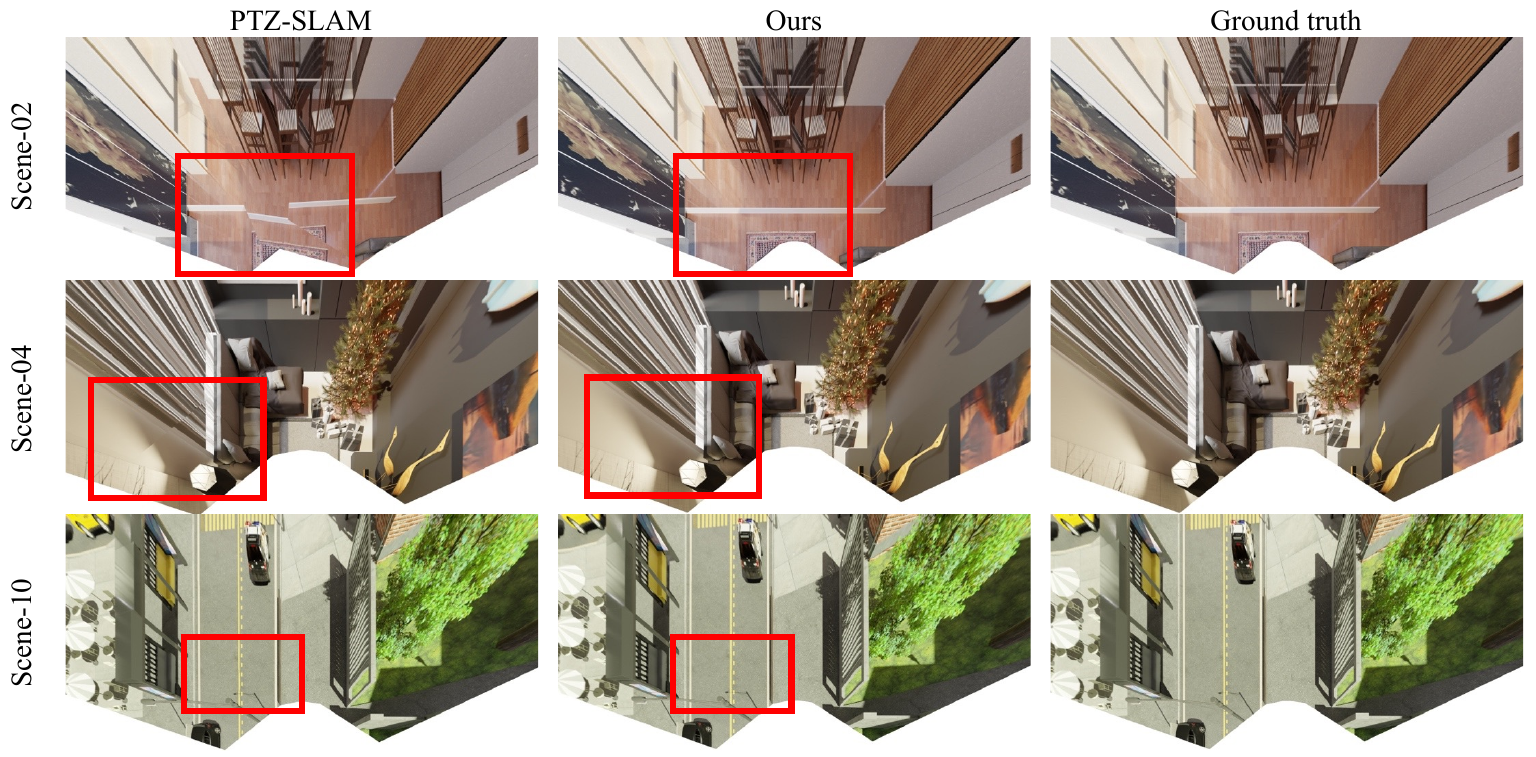}
    \caption{\textbf{Qualitative comparison on the synthetic dataset.} The red boxes indicated the regions to focus on.}
    \label{fig:bev-180-compare}
\end{figure}

\textbf{Synthetic dataset.}
We first compared our method with PTZ-SLAM~\cite{PTZ-SLAM} on synthetic datasets. Since the open-source version of PTZ-SLAM only supports a limited range of pan angles and its running speed has not been optimized, for each scene we selected 10 images for comparison within a 180-degree pan angle range, with the first frame annotated. The camera parameters of the first frame required by PTZ-SLAM were calculated by minimizing the 2D-3D reprojection error using the method described in \cite{hartley2003MVG}.
Results in Table~\ref{tab:Quantitative_Synthetic_Comp} indicts our better performance. Qualitatively, our projections into BEV in Fig.~\ref{fig:bev-180-compare}, align more closely with the ground truth, without notable misalignment at the stitching edges.

\begin{table}[h]
\caption{Comparisons on the Synthetic dataset (PTZ-SLAM vs. Ours).}
\label{tab:Quantitative_Synthetic_Comp}
\centering
    \resizebox{0.95\linewidth}{!}{%
    \begin{tabular}{c c c c c c c}
       \hline
       \multirow{2}{*}{Scenes} & \multicolumn{2}{c}{$FLE(pix) \downarrow$} & \multicolumn{2}{c}{$APE_{rot}(\deg) \downarrow$} & \multicolumn{2}{c}{$APE_{trans}(m)\downarrow$} \\
       \cline{2-7}
       & mean & median & mean & median & mean & median \\
        \hline
        Scene-01 & 3.49, \textbf{0.34} & 3.45, \textbf{0.28} & 0.13, \textbf{0.10} & 0.12, \textbf{0.10} & 0.21, \textbf{0.08} & 0.21, \textbf{0.08} \\
        Scene-02 & 141.49, \textbf{2.49} & 12.42, \textbf{2.58} & 2.36, \textbf{0.10} & 0.21, \textbf{0.09} & 0.32, \textbf{0.14} & 0.32, \textbf{0.14} \\
        Scene-03 & 5.67, \textbf{1.59} & 5.60, \textbf{1.51} & 0.20, \textbf{0.12} & 0.16, \textbf{0.12} & 0.40, \textbf{0.02} & 0.40, \textbf{0.02} \\
        Scene-04 & 21.23, \textbf{1.36} & 11.67, \textbf{1.35} & 0.43, \textbf{0.10} & 0.27, \textbf{0.10} & 0.32, \textbf{0.05} & 0.32, \textbf{0.05} \\
        Scene-05 & \textbf{1.14}, 4.36 & \textbf{0.57}, 4.15 & 0.20, \textbf{0.11} & 0.20, \textbf{0.11} & \textbf{0.14}, \textbf{0.14} & \textbf{0.14}, \textbf{0.14} \\
        Scene-06 & \textbf{0.75}, 1.60 & \textbf{0.78}, 1.65 & \textbf{0.04}, 0.06 & \textbf{0.04}, 0.05 & \textbf{0.06}, 0.10 & \textbf{0.06}, 0.10 \\
        Scene-07 & 7.25, \textbf{3.20} & 6.86, \textbf{2.76} & 0.17, \textbf{0.09} & 0.15, \textbf{0.08} & 0.27, \textbf{0.17} & 0.27, \textbf{0.17} \\
        Scene-08 & 7.82, \textbf{2.44} & 7.68, \textbf{2.41} & \textbf{0.24}, 0.25 & 0.26, \textbf{0.24} & 0.23, \textbf{0.15} & 0.23, \textbf{0.15} \\
        Scene-09 & 6.74, \textbf{1.60} & 6.62, \textbf{1.61} & \textbf{0.15}, 0.17 & \textbf{0.13}, 0.17 & 0.32, \textbf{0.13} & 0.32, \textbf{0.13} \\
        Scene-10 & 6.93, \textbf{2.27} & 6.75, \textbf{2.35} & 0.21, \textbf{0.14} & 0.19, \textbf{0.14} & 0.28, \textbf{0.16} & 0.28, \textbf{0.16} \\
        \hline
        All & 20.25, \textbf{2.13} & 6.16, \textbf{1.86} & 0.41, \textbf{0.12} & 0.16, \textbf{0.11} & 0.26, \textbf{0.11} & 0.28, \textbf{0.13} \\
        \hline
    \end{tabular}%
    }
\end{table}

Then we uniformly selected 30 images panning 360-degree per scene for offline evaluation, with 2 images annotated. We tested the online stage using the remaining 150 images per scene. In Table~\ref{tab:Quantitative_Synthetic_both}, our method consistently achieved high-precision  calibration results in both stages, using only a limited number of annotations. Besides, we add random distortions to test our estimation of distortion coefficients. In Fig.~\ref{fig:all-undistort} (right), the originals show noticeable barrel distortion. Our method effectively corrects this, accurately restoring the geometry and ensuring the straight lines.

\subsection{Performance}

Time consumption for both offline and online stages is detailed in Table~\ref{tab:time_analysis}. The offline stage involves processing 30 images, while the online stage is averaged per image. Efficiency can be increased by reducing reference frames, as most time is spent on feature extraction and matching.

\begin{table}[h]
\caption{Statistics of calibration error on the synthetic dataset.}
\label{tab:Quantitative_Synthetic_both}
    \centering
    \resizebox{0.95\linewidth}{!}{%
    \begin{tabular}{ccccccc|cccccc}
    \hline
    \multirow{3}{*}{Scenes} & \multicolumn{6}{c|}{Offline Stage} & \multicolumn{6}{c}{Online Stage} \\ \cline{2-13} 
     & \multicolumn{2}{c}{$FLE(pix) \downarrow$} & \multicolumn{2}{c}{$APE_{rot}(\deg) \downarrow$} & \multicolumn{2}{c|}{$APE_{trans}(m)\downarrow$} & \multicolumn{2}{c}{$FLE(pix) \downarrow$} & \multicolumn{2}{c}{$APE_{rot}(\deg) \downarrow$} & \multicolumn{2}{c}{$APE_{trans}(m)\downarrow$} \\ \cline{2-13} 
     & mean & median & mean & median & mean & median & mean & median & mean & median & mean & median \\ \hline
    Scene-01 & 1.07 & 0.69 & 0.09 & 0.08 & 0.02 & 0.02 & 1.28 & 0.84 & 0.10 & 0.09 & 0.02 & 0.02 \\
    Scene-02 & 2.34 & 1.96 & 0.17 & 0.15 & 0.14 & 0.14 & 2.62 & 1.96 & 0.21 & 0.17 & 0.14 & 0.14 \\
    Scene-03 & 0.42 & 0.36 & 0.16 & 0.16 & 0.11 & 0.11 & 0.59 & 0.44 & 0.17 & 0.16 & 0.11 & 0.11 \\
    Scene-04 & 1.73 & 1.35 & 0.14 & 0.13 & 0.03 & 0.03 & 1.66 & 1.29 & 0.15 & 0.14 & 0.03 & 0.03 \\
    Scene-05 & 4.86 & 4.94 & 0.12 & 0.13 & 0.10 & 0.10 & 5.12 & 4.97 & 0.14 & 0.13 & 0.10 & 0.10 \\
    Scene-06 & 0.94 & 0.71 & 0.12 & 0.12 & 0.07 & 0.07 & 1.15 & 0.86 & 0.11 & 0.10 & 0.07 & 0.07 \\
    Scene-07 & 1.37 & 1.02 & 0.11 & 0.10 & 0.11 & 0.11 & 1.50 & 1.42 & 0.13 & 0.11 & 0.11 & 0.11 \\
    Scene-08 & 1.18 & 1.15 & 0.06 & 0.06 & 0.10 & 0.10 & 1.14 & 0.95 & 0.08 & 0.07 & 0.10 & 0.10 \\
    Scene-09 & 0.48 & 0.28 & 0.10 & 0.11 & 0.07 & 0.07 & 0.69 & 0.51 & 0.11 & 0.11 & 0.07 & 0.07 \\
    Scene-10 & 0.77 & 0.72 & 0.08 & 0.09 & 0.04 & 0.04 & 0.75 & 0.65 & 0.09 & 0.09 & 0.04 & 0.04 \\
    \hline
    All & 1.52 & 0.97 & 0.12 & 0.11 & 0.08 & 0.08 & 1.65 & 1.02 & 0.13 & 0.12 & 0.08 & 0.08 \\ \hline
    \end{tabular}%
    }
\end{table}

\begin{table}[h]
\caption{Computational time of our method (in seconds).}
\label{tab:time_analysis}
    \centering
    \scalebox{0.85}{
    \begin{tabular}{c c c c c}
       \hline
        & Feature extraction & Feature matching & \parbox{1.5cm}{Non-linear optimization} & Total \\
       \hline
       Offline stage & 2.54 & 8.81 & 11.86 & 29.80 \\
       Online stage & 0.06 & 0.34 & 0.05 & 0.68 \\
        \hline
    \end{tabular}
    }
\end{table}

\subsection{Ablation study}

\textbf{Feature Extraction and Matching.}
We evaluated the impact of different feature extraction and matching algorithms on the offline stage by testing ORB~\cite{orb}, SIFT~\cite{Lowe2004DistinctiveIF}, and KNN feature matching, as shown in Table \ref{tab:ablation_matching}. ORB features often resulted in higher mismatches and lower performance. Although SIFT features provided higher sub-pixel level precision, slightly outperformed SuperPoint and SuperGlue, it is sensitive to illumination and weather changes. We will keep feature matching as a pluggable module for adaptability.

\begin{table}[t!]
\caption{Comparisons of different feature matching methods.}
\label{tab:ablation_matching}
    \centering
    \scalebox{0.9}{
    \begin{tabular}{c c c c c c c}
       \hline
       \multirow{2}{*}{Scenes} & \multicolumn{2}{c}{$FLE(pix) \downarrow$} & \multicolumn{2}{c}{$APE_{rot}(\deg) \downarrow$} & \multicolumn{2}{c}{$APE_{trans}(m)\downarrow$} \\
       \cline{2-7}
       & mean & median & mean & median & mean & median \\
        \hline
        Ours+ORB & 16.48 & 1.16 & 2.42 & 0.17 & 0.27 & \textbf{0.06} \\
        Ours+SIFT & \textbf{0.80} & \textbf{0.32} & \textbf{0.12} & 0.12 & \textbf{0.08} & 0.07 \\
        Ours+SP-SG & 1.52 & 0.97 & \textbf{0.12} & \textbf{0.11} & \textbf{0.08} & 0.08 \\
        \hline
    \end{tabular}
    }
\end{table}

\subsection{Applications}

\begin{figure}[!t]
    \centering
    \includegraphics[width=\linewidth]{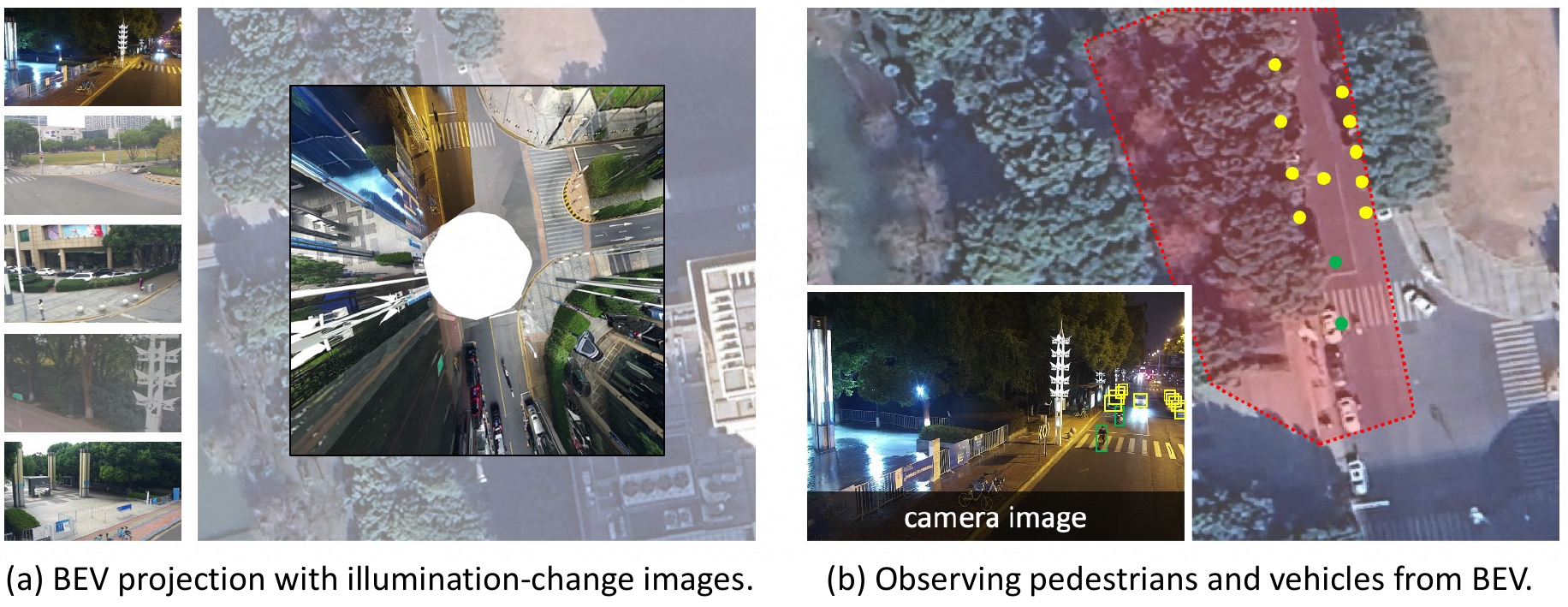}
    \caption{The application in traffic situation awareness.}
    \label{fig:application}
\end{figure}

One application is to apply the method on the PTZ camera at the crossroad and project the detected objects onto the map to understand the traffic condition. As illustrated in Fig.~\ref{fig:application} (a), our algorithm can effectively calibrate images with different illuminations. Specifically, Fig.~\ref{fig:application} (b) demonstrates the precise mapping of pedestrians and vehicles onto Google Maps, achieved through the estimated camera parameters. Additionally, the camera's viewshed, or the area visible from its position, is estimated and depicted in red, further enhancing situational awareness.

\section{Conclusion and future work}

We proposed a two-stage PTZ camera calibration method that can efficiently and accurately estimate camera parameters for arbitrary viewpoints. The introduced PTZ-IBA algorithm provides the underlying theoretical support, thereby enhancing the robustness. Extensive evaluations demonstrate our robustness and superior performance over state-of-the-art methods on various real and synthetic datasets. 
In the future, we plan to support the calibration of PTZ cameras with rotation center offset. Additionally, we intend to decrease annotation requirements even more and explore the viability of adopting an unsupervised approach.

\bibliographystyle{IEEEtran}
\bibliography{IEEEabrv,mybibfile}

\end{document}